# Worsening Perception: Real-time Degradation of Autonomous Vehicle Perception Performance for Simulation of Adverse Weather Conditions

Ivan Fursa[a], Elias Fandi[a], Valentina Muşat[b], Jacob Culley[a], Enric Gil[a], Izzeddin Teeti[a], Louise Bilous[a], Isaac Vander Sluis[c], Alexander Rast[a] and Andrew Bradley[a*]

[a]*Autonomous Driving & Intelligent Transport Group, Oxford Brookes University*
[b]*Oxford Robotics Institute, Oxford University*
[c]*Streetdrone Limited*

**Abstract**

*Autonomous vehicles rely heavily upon their perception subsystems to 'see' the environment in which they operate. Unfortunately, the effect of variable weather conditions presents a significant challenge to object detection algorithms, and thus it is imperative to test the vehicle extensively in all conditions which it may experience. However, development of robust autonomous vehicle subsystems requires repeatable, controlled testing - while real weather is unpredictable and cannot be scheduled. Real-world testing in adverse conditions is an expensive and time-consuming task, often requiring access to specialist facilities. Simulation is commonly relied upon as a substitute, with increasingly visually realistic representations of the real-world being developed. In the context of the complete autonomous vehicle control pipeline, subsystems downstream of perception need to be tested with accurate recreations of the perception system output, rather than focusing on subjective visual realism of the input - whether in simulation or the real world. This study develops the untapped potential of a lightweight weather augmentation method in an autonomous racing vehicle - focusing not on visual accuracy, but rather the effect upon perception subsystem performance in real time. With minimal adjustment, the prototype developed in this study can replicate the effects of water droplets on the camera lens, and fading light conditions. This approach introduces a latency of less than 8 ms using compute hardware well suited to being carried in the vehicle - rendering it ideal for real-time implementation that can be run during experiments in simulation, and augmented reality testing in the real world.*

---

[*] *Corresponding author. E-mail: abradley@brookes.ac.uk*

# 1 Introduction

Autonomous driving presents a number of outstanding challenges for the vehicle to be truly independent. In particular, a crucial part of the overall system is perception - the method by which the Autonomous Vehicle (AV) 'sees' and interprets its environment. The majority of both race and road AVs operate via a pipeline of subsystems responsible for perception, localisation and mapping, decision making, and control (Van Brummelen *et al*., 2018). This creates a complex, deep chain of dependent interacting subsystems, which are ultimately reliant upon the quality of information received from perception. It is not just the perception front-end, but these downstream subsystems that need to be tested rigorously to ensure that they are robust in a wide range of operating conditions. Autonomous racing provides a platform to test each of these subsystems thoroughly, in a sandboxed environment - requiring software which must run in real-time on lightweight hardware (Betz *et al.,* 2019), and intensifying the impact even of marginal losses in performance as conditions vary.

Various weather conditions can significantly degrade perception, and if not accounted for, subsequent subsystems in the AV control system pipeline may be susceptible to incomplete or erroneous information. Weather conditions have historically proven to be one of the more challenging aspects to mitigate against in perception subsystems (Yoneda *et al*., 2019). A vehicle that works well in initial testing may perform poorly in the real world, thus "it is of utmost importance to test automated driving function under adverse weather conditions" (Hasirlioglu & Riener, 2018). In road-going AVs, reducing speed, or even relinquishing control to a human driver in difficult conditions may be options for compensating a performance loss (Ishigooka *et al*., 2019) - but this is fundamentally opposed to the objectives of autonomous racing. In either application, AV technology needs to be developed to be suitable for - and undergo rigorous testing in - the complete variety of environments and weather conditions which may occur.

Two complementary approaches have been used to test AVs. The first, real-world testing, is typically considered the 'gold standard'. However, real-world development and testing of AVs is expensive, inherently risky (thus typically requires access to a specialised testing facility), and demands many hours - or indeed months - of data collection and iterative improvements. Data gathering requires hours of driving, under different sets of conditions, which of course themselves cannot be scheduled: the data collection process is at the mercy of what the weather decides to do. The second approach, simulation, thus offers a number of appealing advantages (Li *et al*., 2019) and can be a valuable additional resource given the time and expense of real-world data collection. Techniques including Software-in-Loop and Hardware-in-Loop testing have been used to replicate real-world failure cases to accelerate development - culminating in complete Vehicle-in-Loop testing in a hybrid environment (Solmaz *et al*., 2021). This suggests a third approach: augmented-reality testing, where additional, virtual effects are layered upon the sensors of a real vehicle operating in the real world.

Rigorous and controlled testing of the vehicle control system in various weather conditions requires the conditions themselves to be repeatable, but real-world weather simply cannot offer any such guarantees. Opportunistic testing when the 'right' weather conditions happen to materialise lacks rigour - thus a way to introduce reliable repeatable weather effects, to both real-world and simulated testing, is imperative.

The typical approach to dealing with adverse weather has been to treat it as a pure perception problem, focused on enhancing performance by using off-line simulated or augmented weather (Halder *et al*., 2019; Hnewa & Radha, 2020). However, such techniques neglect the performance and testing requirements of downstream (*e.g.* mapping) subsystems, and cannot be applied in real time - where what is happening in the moment affects how the vehicle responds. In simulation, it might be thought appropriate to test off-line or, *e.g.* by slowing the update rate, but this is not a rigorous test of the real-time system capabilities, and in any case cannot be applied to real-world testing. Thus is it desirable, indeed arguably essential, for the downstream effects of weather degradations to be introduced on-line, while the AV is actually in operation (whether in the real world or in simulation).

An intriguing approach is to introduce real-time effects to improve perception in adverse weather conditions using online 'de-weathering' (Hassan *et al.*, 2020) - however there are very few studies which deliberately attempt to *worsen* the perception of an autonomous vehicle in the real world - let alone an autonomous racing vehicle. Autonomous racing provides the perfect test-bed to explore the impact of heavily impaired perception, since the intrinsic dangers of testing in real-world conditions can be mitigated using controlled test environments with strict safety protocols, and where AV control systems are pushed to their performance limits.

Ultimately, developing robust control systems relies precisely upon repeatable testing in non-ideal conditions, which cannot themselves be relied upon. A simple, easily-adjustable, straightforward method of generating the effects of adverse weather and applying them to the control system pipeline, without introducing high additional computing load, would be highly desirable to enable such repeatability. This paper explores the potential of applying 'drop-in' real-time augmentations to actively worsen perception subsystem performance as part of a pipeline for both simulation and real-world testing in an autonomous racing vehicle.

## 2 Related work

This study represents the first attempt to introduce online degradation of perception in a complete system pipeline for autonomous racing. A typical AV control system is composed of several distinct components implemented as follows. The perception subsystem commonly employs an AI object detector to identify particular landmarks of interest (*e.g.* road signs, traffic lights *etc.*). The perception pipeline also includes detection of transient objects, (*e.g.* pedestrians and other road users), typically using some sort of machine learning-based object identification (Pendleton *et al.*, 2017). The location and range of the landmarks from the vehicle's perspective are then passed to a localisation and mapping subsystem - which generates an updated map based on the detected position of landmarks, and identifies the location and orientation of the AV within this map (Jo *et al.*, 2018). The information regarding the map and other objects is then used by a decision subsystem to select how to proceed, and plan a trajectory through the environment - which is passed to the control subsystem that generates commands to drive the vehicle physically. Testing of such subsystems requires consideration of the impact of weather on each stage of the pipeline.

For real-world testing in adverse weather, specialist facilities (Bijelic *et al.*, 2018) exist that can physically emulate various weather conditions, but these are expensive, need to be carefully scheduled, and are not typically accessible to all groups who might wish to use them. Some simulators (Dosovitskiy *et al.*, 2017; Shah *et al.*, 2018; Best *et al.*, 2018) have made efforts to incorporate the physical and visual effects of weather (Rosique, *et al.* 2019), *e.g.* modelling the effect of wet roads on tyre behaviour, and modelling visual effects representing various weather conditions, *e.g.* snow. However, the focus of such simulators with respect to perception is typically to model the impact of accurate weather effects on the sensor 'front-end' rather than the perception 'back-end'.

## 2.1 Visually realistic weathering

There have been several approaches to recreating weather effects artificially on real-world imagery. A popular method is to employ graphics-editing tools from existing open-source frameworks such as OpenGL (Praveen *et al.*, 2017) to apply effects that simulate weather conditions. Models have been developed to simulate the rain by modelling the traversing of light through raindrops (Bernard *et al.*, 2013), while Creus & Patow (2013) developed a photorealistic rendering of rain by modelling the dynamics of the raindrops. However, these engine-based approaches use sophisticated physics models - thus are computationally expensive, and require specialist knowledge of parameters specific to the conditions. Typically, these parameters are empirically determined by subjective assessment of the realism of the generated images. It is therefore an open question how accurately such recreations model the degradation of real sensor performance in adverse weather conditions.

Recent advances in machine learning approaches to image processing have led to the development of methods that enable both weather appearance transfer and generation - with many approaches aimed at weather removal (Yang *et al*, 2020). Generative Adversarial Networks (and their variants) have been successfully used for de-weathering (*i.e.* de-raining, de-hazing) - for example Zhang *et al.* (2020), Porav *et al.* (2019), Porav *et al.* (2020), Anvari *et al.* (2020) and Uřičář *et al.* (2019), whereas weathering (*i.e.* adding fog) and light manipulation is done by Lin *et al.* (2019), Porav *et al.* (2018) and Dai *et al.* (2020). Whilst these AI approaches offer visually pleasing weather effects, they generally require hardware acceleration (GPUs), and extensive training datasets in each target weather condition. As a result, the possible weather effects are limited by the distribution of conditions exhibited in the training data - and thus introducing degradations of a class or severity not encountered in the original dataset, in a controllable manner, is a formidable undertaking. Generally this will create significant development effort and require further (offline) training with large datasets. As such, most of the work has been applied in single weather conditions, in offline urban or more general scenes, as opposed to the real-time conditions of autonomous racing.

Though visually realistic, simulating the physics directly is computationally expensive and comes with no strong guarantee that it is better at modelling the downstream effect. Meanwhile, the machine learning approaches require extensive training data. Thus there is a need for a simpler, computationally cheaper methodology for simulating adverse weather, attuned to the downstream processing requirements. Such approaches could include noise addition, rotations, translations, random cropping and flipping (Shorten *et al.*, 2019). In this context, Rusak *et al.* (2020) explore Gaussian noise and a learned Adversarial noise to make neural networks more robust, while Moreno-Barea *et al.* (2018) explore both Uniform and Gaussian noise. Other effects, such as rain and fog, have been applied previously, with an emphasis on photorealism. For example, Volk *et al.* (2019) used classic augmentations including Gaussian noise or Salt-and-Pepper noise as a benchmark with which to compare their more complex physics-based rendering. These augmentations are commonly used in offline settings, such as for neural network training (to enhance datasets), but fewer uses have been explored for online augmentation - although the potential has existed for some time (Hernandez-Lopez & Rivera, 2015).

## 2.2 Modelling weather effects upon sensor & perception performance

In contrast to the studies above which largely focus on weather photo-realism, this study considers the impact of weather as a real-time component affecting an integrated perception, mapping, and control pipeline. In such a pipeline, where components are connected through interfaces that abstract the functionality of other components, downstream subsystems are unaware of the source of their data. It is largely irrelevant how the data is generated or how realistic the imagery is: it only matters that the data produces similar behaviour in the systems concerned. In other words, it is not essential that the weather in the simulated images *looks* realistic; it is much more important that the effects of the modelled weather result in a realistic change in the performance of the perception subsystem.

To this end, Hasirlioglu & Riener (2018) develop noise filters for camera, LiDAR and Radar sensors and compare them with rain streaks simulated in a test facility where the rain parameters (*i.e.* rain intensity, drop size distribution) can be controlled. Although initially the validation of the effects is done at sensor level, in a follow up paper (Hasirlioglu & Riener, 2020) they test object detection (using *YOLOv3*) as a function of the distance between the sensor and the object to be detected. Although the study highlights the opportunity to explore the use of similar modelling on simulated datasets, they present limited results in terms of applying the rain effects upon synthetic imagery.

A framework to simulate and analyse noise factors was proposed by Chan *et al*. (2020) to predict sensor performance degradation in autonomous vehicles, concluding that analysis of noise factors individually is insufficient, and as a result perception is drastically affected when multiple noise sources are combined. However, this study focuses upon LiDAR data rather than visual imagery. In a similar vein, Byeon & Yoon (2020) simulate the effects of rain on synthetic sensory data, again focussing on LiDAR sensors.

Halder *et al*. (2019) propose a realistic physics-based rain and fog rendering, quantifying the resulting reduction in perception performance, and performing both quantitative and qualitative analysis of the realism of the images. However, as with several of the studies in Section 2.1, the application is in offline training data generation for enhancing object detector performance, rather than the deliberate, online worsening of perception to enable development of the downstream tasks in the autonomous control system pipeline.

## 2.3 Weather modelling in autonomous racing

Autonomous racing demands rapid design cycles to remain competitive - motivating studies exploring real-time weather modelling solutions to exploit performance gains in challenging conditions, where low-cost, repeatable testing is essential. Simulation has been the main focus thus far. For example, Zadok *et al*. (2019) explore conditions including light intensity and cloudiness, distortions and lens flare reflection by modelling these effects directly into the simulation environment using *Airsim*. They achieve impressive visual representations which enabled them to train a vehicle to drive in the real world using end-to-end learning in a simulator. However, in the end-to-end approach, there is no introspection into the pipeline used for controlling the vehicle - and thus no way to measure, validate or quantitatively worsen the performance of a perception subsystem in response to weather conditions. Furthermore, this implementation of weathering effects is built into the simulation, and thus cannot be mapped into real-world testing.

Culley *et al.* (2020) developed a simulation platform which is interchangeable with the real autonomous racing vehicle, proposing the use of simple noise models upon the simulated camera images and other sensors to worsen the perception subsystem performance artificially in simulation. Similarly to Zadok *et al*. (2019), these (very basic) models were only applicable in simulation, not real-world testing. They were implemented with no specific weather condition in mind, and no validation against real-world perception performance was performed. This prior work formed the basis of the system which is utilised and further developed in this study, and will be discussed in more detail in Section 3.2.

# 3 Methodology

## 3.1 Method overview

The downstream processing algorithms for localisation, mapping, path planning and control are ignorant of the aesthetic appearance of the images produced by the camera - they simply receive identified objects, with range and centreline offset information. If the perception subsystem is unable to identify an object accurately, downstream algorithms will receive incomplete or erroneous information and may cause location and map to diverge from reality.

Thus, the methodology developed in this study aims to create a pipeline of weather augmentations that can be run in real time. Rather than focussing upon accurately modelling the visual quality of weather effects, this study focuses on developing a model which accurately replicates the perception performance found in real-world weather conditions. The model is designed to be lightweight, introduce minimal latency, and run using hardware suitable for carrying on-board in a racing vehicle to facilitate application to both simulation and real-world testing. The technique can be run in online mode as part of the pipeline, or offline to post-process existing data into a range of identical scenes in different weather conditions.

The following sections describe the simulation environment used for this study, the perception subsystem in operation on the vehicle, and the method of implementing the weather augmentations.

Both pathways (real and simulated) feed the perception pipeline, which forms the input for the localisation and mapping subsystems (not shown). The perception pipeline can receive either unmodified data from a simulated or real vehicle, or data with weather augmentation superimposed. A separate inference comparison acts as a test harness and records the difference in object detection accuracy between real and augmented inputs.

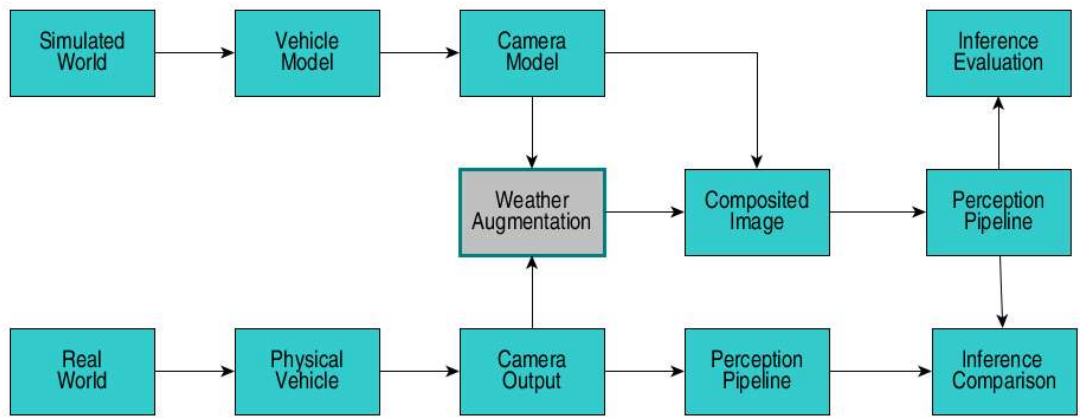

*Figure 1: Weather augmentations can be applied to both real and simulated camera feeds*

## 3.2. The software pipeline

This section describes the software setup and perception pipeline as illustrated in Figure 1. The majority of the software is a development of the work of Culley *et al.* (2020), comprising an autonomous control system with an open-source object detector for perception, which passes the detected objects to a pipeline of subsystems responsible for mapping, decision making and vehicle control. The AV control system can be used on the real vehicle, or coupled with a tightly integrated simulation which mimics the real vehicle. This software, and the developments outlined below were built in Python 3.8 (Python Software Foundation, 2019) utilising ROS2-Foxy (Open Robotics, 2020) middleware layer to connect modules.

### *3.2.1 Replicating the real world*

For testing, a number of real-world datasets were generated under various conditions. Simple circuits were laid out using cones (Figure 2) and simulated replicas of the real world testing scenarios were developed in the Gazebo simulator (Open Robotics, 2020) based upon manual measurement. Care was taken to replicate accurately the positions of the objects to be identified by the perception subsystem (in this case, cones) demarcating the boundaries of the test track, such that the structure is kept consistent and does not play a role in the variation of performance. The methods of Ahamed *et al*. (2018), which simply modelled the circuit, were extended to provide a representation of the actual environment encountered by the vehicle and perception subsystem, including salient surrounding objects and features.

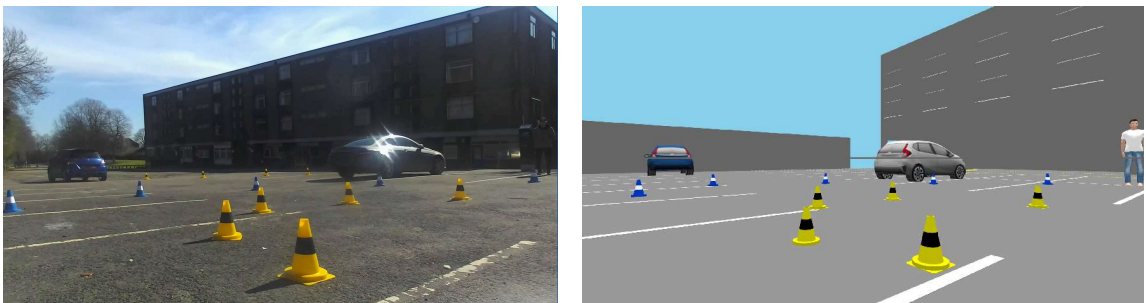

*Figure 2: Test track layout (left) and surrounding environment modelled in simulation (right)*

### *3.2.2 Simulating the vehicle and sensors*

The simulation described by Culley *et al.* (2020) was selected as it incorporates accurate models of the vehicle dynamic behaviour, CAN communications, and all sensors from the real autonomous vehicle - including GPS, IMU, stereoscopic cameras, and LiDAR sensors. The simulation was run on an Intel 16-core Xeon processor and an Nvidia GTX 980 graphics card.

Since the focus of this study is on modelling the effects of weather upon vision-based sensors, the camera models in Gazebo were developed further to replicate the real-world camera feed accurately, including parameters defining the mounting position and angle of the camera upon the vehicle, horizontal / vertical field of view, CCD sensor size, focal length and frame rate. The simple camera noise model described by Culley *et al*. (2020) was removed from the simulation, so that weather effects could be inserted as a post-process using the augmentations presented in the following sections.

### 3.2.3 Perception subsystem

The perception subsystem uses YOLOv3-tiny (Redmon and Fahradi 2018) to identify the objects of interest (in this case, cones) from the camera input. This architecture efficiently processes images at frame rates which are orders of magnitude faster than commonly-employed algorithms such as R-CNN and SSD (Adarsh *et al.*, 2020) - making it ideally suited to the high speed, real-time demands of autonomous racing using lightweight compute hardware. Processing was undertaken at a resolution of 608×352, providing a balance between frame rate and object detection accuracy. The depth map from the camera is then used to provide visual range indication, which is combined with the detected objects to build a map of cone locations in the body reference frame of the vehicle. This (local) map provides the input to the downstream localisation and (global) mapping subsystem.

The perception subsystem was developed and trained in Culley *et al.* (2020), using more than 1000 real-world images (none from simulation) containing over 10,000 manually labelled cones from data captured from laps of a circuit. A variety of standard augmentations were performed upon the images during training, including flipping, scaling, tiling and HSV adjustments to expand the dataset.

### 3.3 Noise models

On the track, light conditions can deteriorate rapidly late in the day, and droplets may adhere to the camera lens if racing is taking place on wet surfaces - whether it is currently raining or not. Therefore, this study prioritises augmentations for low light conditions and water droplets on camera lenses.

The weather effects were replicated by employing image augmentation and deterioration using a modified version of the *Albumentations* library (Buslaev *et al.*, 2020) in PyTorch (Paszke *et al.*, 2019), and Pillow - a basic image processing library (Pillow, 2021). *Albumentations* was selected as it contains a plethora of additional weather effects (such as rain, fog, shadow, snow, noise, histogram matching), allows for composable augmentations, and has been optimised for fast performance - making it ideally suited to real-time application in an autonomous racing vehicle.

It should be noted that the simulation of adherent droplets was achieved by adapting the 'Fog' effect, as it was empirically found to replicate them well visually in terms of randomness, transparency and size. Since Fog is a composite effect combining both droplets and blurring, the library was modified to isolate the droplet effect by removing the blurring. The degree of degradation was controlled by varying the *alpha* parameter, with the *fog coefficient* kept constant (Table 1).

| Weather effect | Description | Control Parameter | Value | Latency[1] (ms) |
|---|---|---|---|---|
| Adherent droplets | Randomly applies droplet effects (circles) onto image, with adjustable transparency | *alpha* | $k_{droplet}$ (Varied between 0-1) | 8 |
| | | *fog_higher_coefficient* | 0.4 | |
| | | *fog_lower_coefficient* | | |
| Light dimming | Reduces the overall brightness of the image | *1-enhancement factor* | $k_{dim}$ (Varied between 0-1) | 3 |

*Table 1: Augmented weather effects, control parameters and latency incurred*

---

[1] *Calculated on an Intel(R) Core(TM) i7-10510U CPU @ 1.80GHz with 16 GB RAM; image resolution 672x376*

Although this study targets an overall worsening of perception performance, it is important to note that each augmentation operates in a very different manner - light dimming is applied uniformly across the entire image, while the droplets exhibit more random behaviour components, such as random positions for circles and randomly varying degrees of transparency. Thus each augmentation has a distinct effect upon the resulting degradation in the perception subsystem performance.

### 3.4 Dataset capture and generation

In order to establish the effects of the augmented weather conditions upon both real and simulated datasets, experimental data was obtained both from real world testing and from a simulated replica of the real-world test scenario. Several sets of keyframes were extracted from each video, and the cones within each keyframe were manually labelled to provide a 'ground truth' - against which the perception subsystem's object detection performance can be compared, in diverse (real and augmented) weather conditions.

#### *3.4.1 Real-world experimental data*

Experimental data was gathered by collecting successive camera data from laps of two different circuits, in a selection of different real-world weather conditions including:

- 'Ideal' conditions (*i.e.* bright but overcast, such that the objects are clearly visible, without excessive shadows or lens flare effects)
- Corresponding 'ideal' conditions with water droplets on the camera lens (manually added with a water spray)
- Low light conditions (taken in the late afternoon and early evening in fading light)

In total, 3058 images were collected from 27 separate real-world datasets, containing a total of 23,990 objects which were then manually labelled. Datasets were manually sorted into 5 subcategories with a subjective evaluation of the droplet severity (Table 2). Categorisation was performed separately by 5 individuals, whose classifications were then averaged. Low light datasets were taken at intervals as the light faded, with classification performed chronologically.

#### *3.4.2 Simulator data*

Datasets are composed of keyframes gathered from the simulator as described in Section 3.2.1. All simulations were performed without additional weather, under uniform lighting resembling bright, overcast weather, and a brightness and contrast adjustment was applied to the simulation output to obtain the highest possible performance from the perception subsystem. 354 images were acquired from the simulator, containing 3280 manually labelled objects (Table 2). Each of the weather effects can then be layered upon each successive frame of the simulator output to produce a composited 'weathered' image.

| Condition | Severity $S$ | Description | Images | Objects | mAP |
|---|---|---|---|---|---|
| Ideal | 0 | Real world, good weather | 224 | 1827 | 0.793 |
| Adherent droplets (Real world) | 1 | Very light droplets, occasional minimal impact upon object visibility | 766 | 5944 | 0.796 |
| | 2 | Light droplets, object visibility occasionally more difficult | 764 | 4950 | 0.763 |
| | 3 | Moderate droplets, object visibility commonly impaired | 372 | 4493 | 0.730 |
| | 4 | Heavy droplets, object visibility frequently severely impaired | 188 | 1197 | 0.587 |
| Low light (Real world) | 1 | Late afternoon | 221 | 1768 | 0.639 |
| | 2 | Sunset | 206 | 1548 | 0.273 |
| | 3 | Dusk | 213 | 1489 | 0.043 |
| | 4 | Night | 104 | 774 | 0.010 |
| Simulated ideal | 0 | Simulated uniform lighting conditions (*i.e.* bright yet overcast) | 354 | 3280 | 0.715 |
| | | **Total** | 3412 | 27270 | |

*Table 2: Testing datasets, subjective classification of condition,
and mean Average Precision (mAP) when run through the perception subsystem*

### *3.4.3 Perception performance analysis*
For the baseline, the pre-trained YOLOv3-tiny model was tested under 'ideal' illumination conditions on both real and simulated data. These conditions yield uniform isotropic lighting with few shadows and high contrast, resulting in accurate detections of objects in the image - measured as a mean average precision (mAP) since the objects of interest - cones in 2 distinct colours - are in multiple, but hierarchical classes (Padilla *et al.*, 2020). This process was repeated for each of the real-world weather conditions, resulting in the measure of perception performance in each severity of condition shown in Table 2. An overview of the complete experimental method is described in the following section.

## 3.5 Experimental method

To facilitate rapid bulk testing, it is desirable to develop a simple parameter selection method for each weather augmentation such that parameters can be easily adjusted to replicate any given 'real world' environmental conditions. The following method describes this process.

1) Data was collected from laps of the circuit in a variety of real-world weather conditions (Section 3.4.1), and perception performance quantified to produce a series of experimental performance figures corresponding to the severity of the relevant condition.

2) The simulator was set up with a track layout as close as possible to the real-world scenario (Section 3.2.1), and data was collected from multiple laps of the circuit. This creates a reasonable basis for comparison of the performance between real world and simulation.

3) The relevant augmentation (adherent droplets or dimming) was performed upon both the 'Ideal' and 'Simulated ideal' datasets, to generate composited images. The relevant critical parameter in each case was varied to control the intensity of the effect. The composited images were then passed through the perception pipeline, with object detection performance quantified at each intensity.

4) Comparisons were made between the perception degradation in both augmented reality and real-world weather conditions to enable identification of parameters which can replicate the effect of various real-world weather severities upon perception performance.

5) Perception performance using augmented simulated data was compared to that from real-world weather conditions, to determine augmentation parameters which can be layered upon the simulator output to replicate the effect of real-world weather in the simulator.

# 4 Results

## 4.1 Adherent droplets

The adherent droplet augmentation provides an effect which generates simple circular occlusions in areas of the image, with varying degrees of transparency. The positions of the occlusions are random, which means that individual object detections within the image are confused randomly: some will be obscured while others may be unaffected. The resulting degradation in perception subsystem performance as the control parameter is varied is shown in Figure 3.

Increasing the strength of the effect produces a result analogous to greater precipitation or moisture in the atmosphere - more droplets accumulate, resulting in a greater degradation in perception performance. Sweeps were run 5 times for each parameter value and averaged to produce the results shown.

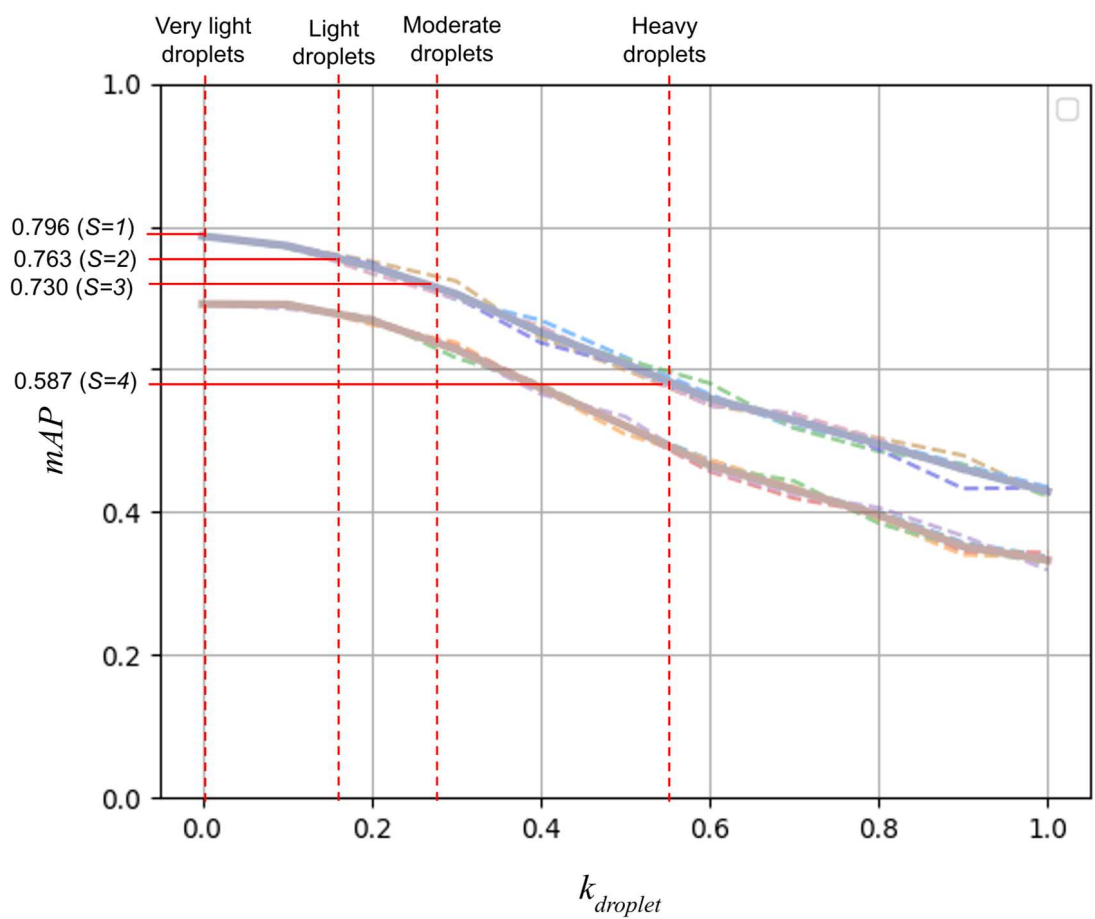

*Figure 3: Effect of adherent droplet augmentation upon perception performance on real-world camera (blue) and simulator output (brown) - Solid line shows the mean of all sweeps. Horizontal and vertical lines indicate the experimental mAP from the corresponding severity (in real-world testing).*

The degradation of perception performance is approximately linear as $k_{droplet}$ is increased, with both simulation and the augmented real-world degrading at the same rate for a given $k_{droplet}$ value. Very light real-world droplets upon the lens were found to have negligible (*+0.003* mAP) effect upon the perception performance, since the impact on visibility is minor and occlusions are rare. The mapping from real-world adverse weather performance enables the identification of a parameter value which can be applied for use in the augmentation, resulting in either an augmented reality, or a simulation in which the object detection is impaired in a comparable manner to that of the real world. Examples are shown in Figure 4, with the corresponding bounding boxes depicting the object detections.

Object detection in the simulated images exhibits a consistent reduction in performance to that of the real-world images - differing by approximately 8 percentage points - which is attributed to the use of a perception model which was trained solely using real-world labelled imagery, and the visual limitations of the Gazebo simulation software.

With further enhancement of the simulation data, it is believed that the parameter could be tuned using either of the two datasets and applied directly to the other to provide similar perception performance in conditions with adherent droplets.

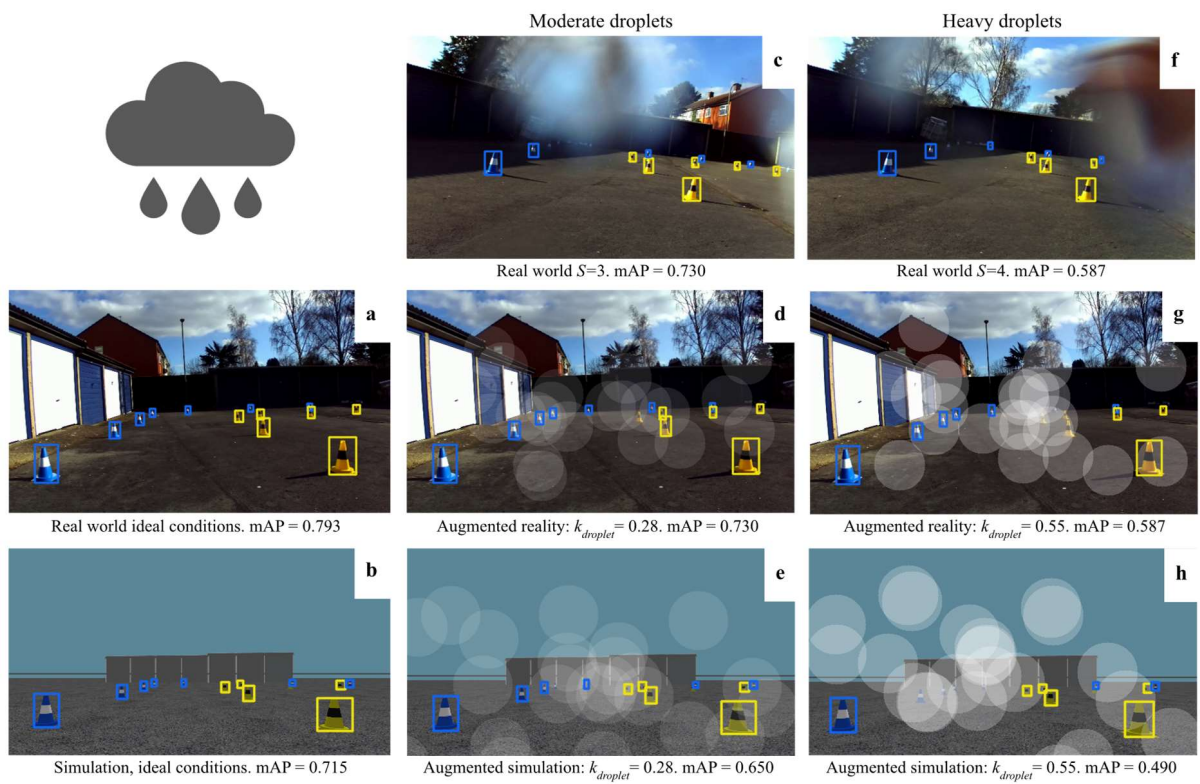

Figure 4: Real and simulated camera imagery with corresponding droplet augmentations, objects detected by the perception system in each case, and mAP for the associated dataset.

## 4.2 Light dimming

In contrast to the random nature of the droplet augmentation, the light dimming effect provides a global worsening of the perception subsystem performance, similar to that experienced in the real world as the light fades in the evening. As stronger effects are applied, the object detector performance drops significantly in both cases (Figure 5).

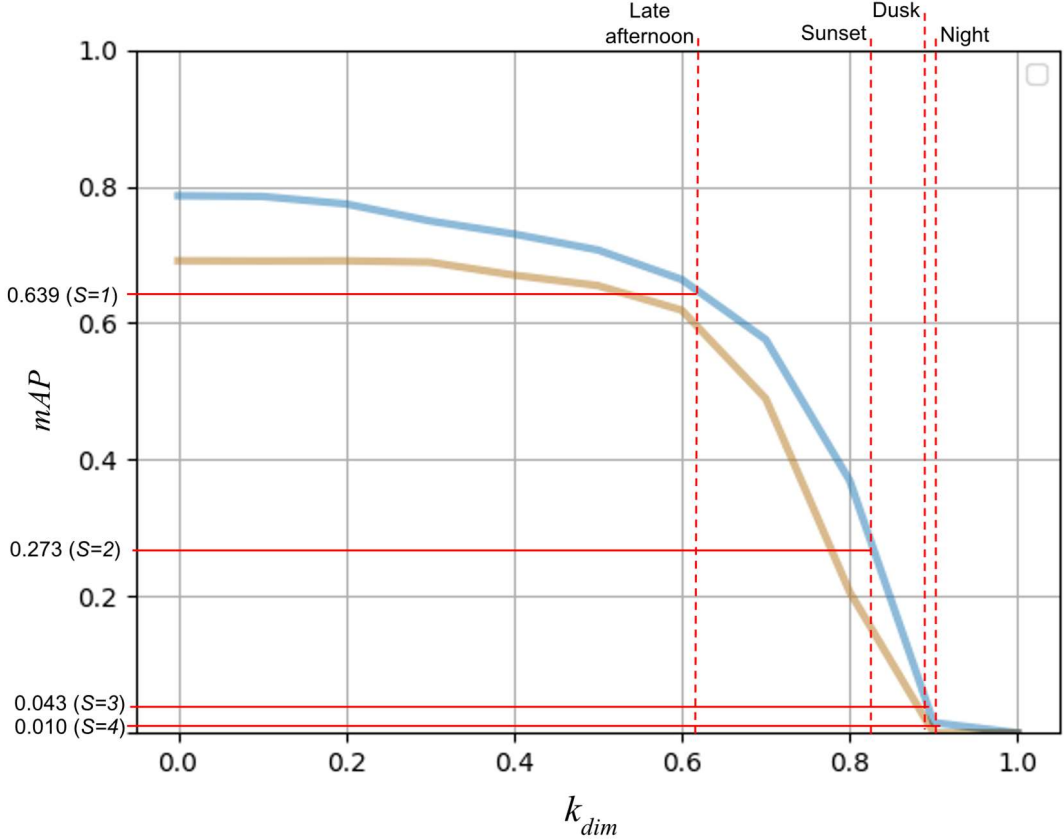

*Figure 5: Effect of light dimming augmentation upon perception performance on real-world camera (blue) and simulator output (brown). Horizontal and vertical lines indicate the experimental mAP from the corresponding severity (in real-world testing).*

The drop in perception performance as the light fades is approximately exponential, in agreement with theory on the effect of illumination on object detection (Pae *et al.*, 2018). Notably, the *Late afternoon* and *Night* datasets were separated by 20 minutes on a winter's day - highlighting the severe consequences of rapidly fading light, and reinforcing the need for careful testing of each subsystem in low light levels. Even at severity *S=2*, perception performance is poor, and is effectively unusable at severity 3 and 4. This is partly due to the perception system having been trained on daylight data - notwithstanding, low light has a severe impact on perception and hence overall system performance.

Examples of the detections on both real-world and simulated data with augmentations are shown in Figure 6. Once again, the $k_{dim}$ parameters were selected based upon the mapping from real dimming data to augmented impaired real-world data. Similarly to the results seen in Section 4.1, the mAP from augmented simulation is approximately 7 percentage points lower than that from augmented reality with a given parameter setting. In moderate lighting conditions, the simulated data can thus usefully serve as a lower bound on expected real-world performance - although as the light fades such a bound loses significance due to poor absolute perception performance in low lighting conditions.

To obtain the best possible accuracy, care should be taken when carrying across the $k_{dim}$ value between real and simulated conditions. Due to directional effects of the lighting and blurring in the real world, the edges of objects from the simulator remain more clearly defined than those from the real camera at very low lighting levels. Additional work may be required to model real-world effects such as blur or the camera automatically adjusting ISO sensitivity for changes in lighting. The use of more advanced simulator scene rendering could mitigate the drop in performance.

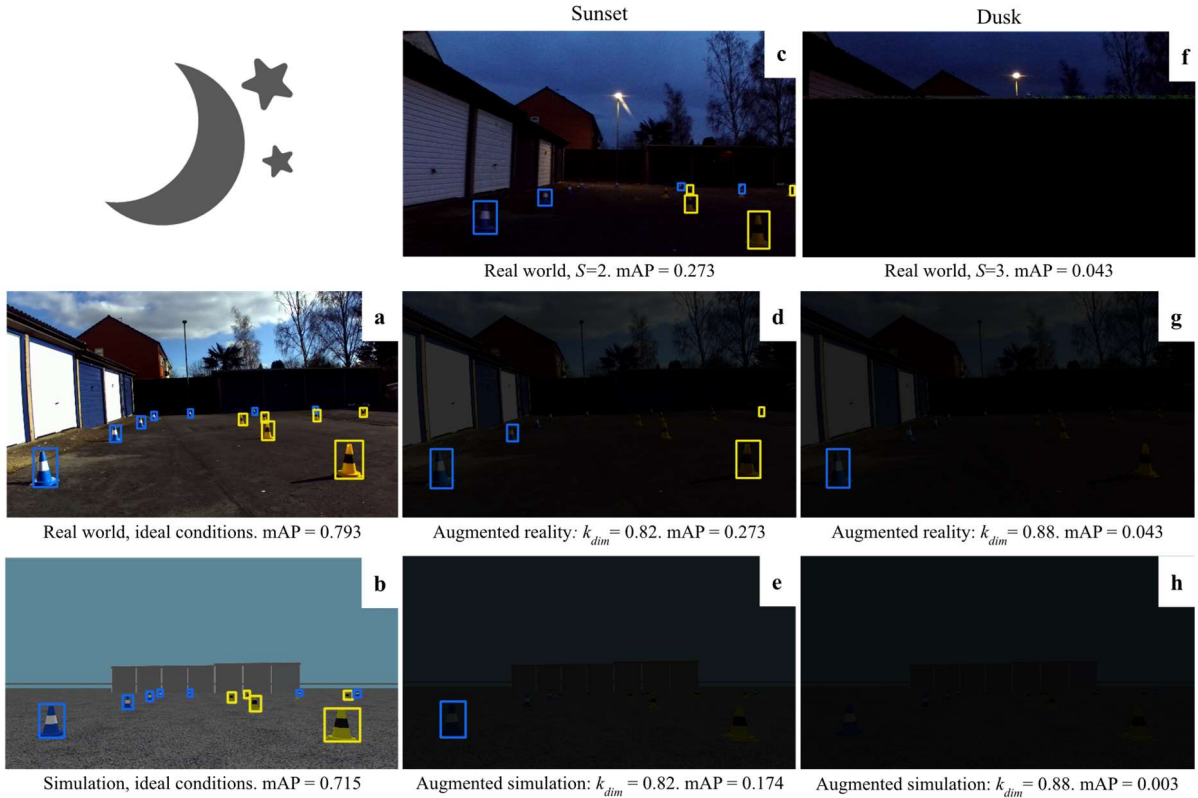

Figure 6: Real and simulated camera imagery with corresponding low light augmentations, objects detected in each case, and mAP for the associated dataset.

Detections are noticeably poor when there is a loss of contrast either from occlusion by droplet aggregations or overall darkening. Cones are often human-perceptible, even where the object detector does not find them (Figure 7). Although outside the scope of the present study, qualitative results suggest some form of saliency map could aid in disambiguating failure cases. This could form part of an attention processing layer concentrating object detection on regions of interest, potentially improving mAP. Such a layer could also be used for offline analysis of cases with poor detection performance, subject to design decisions concerning the saliency algorithm and parameters.

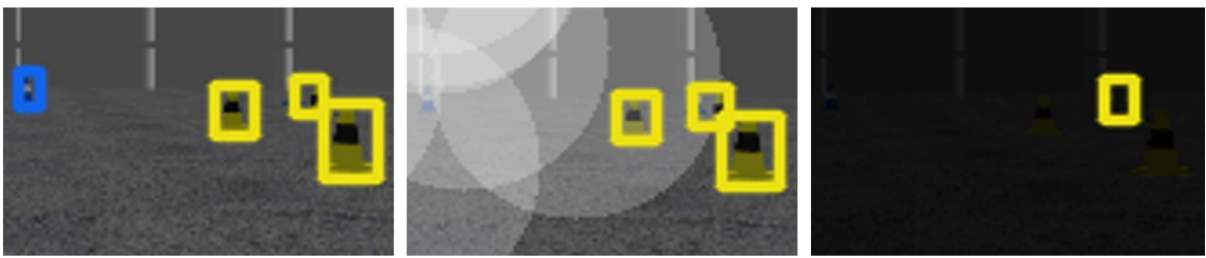

Figure 7: Simulation (left) mAP = 0.715, with corresponding augmentations $k_{droplet}$ = 0.55 (centre) yielding mAP = 0.490, and $k_{dim}$ = 0.78 (right) yielding mAP = 0.273

# 5 Discussion

This study has indicated both the need and the potential to use simple and easily tunable image degradation methods to emulate the real-world effects of weather upon perception subsystem performance. These effects can be applied as a post-process, thus existing datasets can be enhanced as easily as full simulations performed *de novo*. Latencies introduced were minimal (~5 ms - Table 1) and effects can be applied in real time without noticeable impact on performance, minimising perception subsystem delay. In turn, this means there is minimal risk of outdated perception data at higher velocities being distributed to other subsystems, which could cause degradation of racing performance. Critical to the method is the observation that the qualitatively perceived 'realism' of the degradation is irrelevant - the goal is to replicate accurately what the downstream subsystems of the entire AV control pipeline will receive.

The effects generated in this study provide a straightforward alternative to more high-fidelity approaches such as physics modelling or style-transfer using machine learning, reducing the computational overhead and development time required. Such effects are low-cost and could be easily applied to autonomous racing and other AV development applications, without requiring investment in complex proprietary weathering pipelines or expensive simulation hardware. The augmentations are applied as a drop-in component to the perception pipeline - and thus can be applied to data from any simulation platform, or to replicate the effects of adverse weather during real-world testing. It could be used on an otherwise un-weatherproofed vehicle, deferring and reducing the need to protect sensitive, potentially expensive on-board equipment from the physical effects of the weather, or conversely to avoid having to wait for the 'right' weather conditions to materialise - thus accelerating development and reducing cost.

Ongoing work is focused upon improving perception performance and consistency across both simulation and real-world environments, along with investigating both the intrinsic effect of weather degradations upon other processing tasks (e.g. depth estimation), and the downstream impact in the overall AV control pipeline. At present, there remains a difference between absolute mAP in real-world *vs.* simulation, since the simulation provides relatively rudimentary visuals - and the perception system was trained on real-world imagery only. Though not insignificant, the difference is approximately constant across degradation levels, and thus provides a useful lower bound of the performance which could be expected. An area of current investigation is an expanded training dataset including simulated images, to enable perception in simulation to better represent the real world. This would allow an improved perception pipeline with higher mAP both in real-world and simulation, and will form the basis of an iterative test suite to be refined and enhanced in further generations of the autonomous control system. With this further work, it is expected that parameter values can be matched between simulation and augmented real-world.

Whilst the focus of this study is not on enhancing the perception system, a major finding is that low lighting conditions have a considerably larger impact upon perception than adherent droplets on the camera lens. The perception subsystem exhibited very poor performance under serious dimming conditions. Preliminary work with datasets including dark images suggests that performance could be improved with more training, yet in worst-case low light scenarios with severity *S=4*, performance remains unusable. Night driving clearly is one of the significant problems to be addressed in autonomous vehicles - and from the results of this study, there are indications that the challenges may have been underestimated. Human perception features two different vision systems for day and night conditions (Morshedian, 2017), and perhaps a similar approach to AV perception, involving separate day and night sensor pipelines, may be worthy of investigation. This study provides results that can aid development of such a system.

Subject to data acquisition, future work will also examine a broader range of conditions, *e.g.* snow, heavy rain, *etc*. Lighting within simulation is an implementation-specific design choice, and ideally should be matched to real-world, good weather conditions. The *Albumentations* library provides the facility to model sun flare and random shadows, which were briefly experimented with - the former of which (in the authors' opinion) is one of the more visually realistic effects in the library. Realistic addition of correctly-placed shadows is a considerably more complex augmentation than the library provides, suggesting that it may be more appropriate to add directional lighting and shadows within the simulation environment itself. Looking beyond (optical) cameras, modern AVs use a suite of sensors including LiDAR and RADAR - each of which may be affected differently by prevailing weather (Zang, 2019). The concept introduced in this study: deliberately worsening individual sensors using simple, easily-deployed methods, could be used in a more extensive multi-modal sensor modelling scenario to accelerate development of robust, weather-tolerant sensor fusion techniques.

Thus far the experimental work in poor weather has been confined to human driven testing to measure perception subsystem performance - in future the enhanced autonomous control pipeline will be required to drive the car physically in a variety of inclement conditions - which will pose additional challenges, in the form of controlling a vehicle where the dynamic behaviour may be very different due to the reduction in grip on wet road surfaces.

Poor perception appears to have been traditionally treated as a problem to be 'solved' or optimised away by training or pre-processing of the data in the perception front-end (Finlayson, 2018; Halder *et al.,* 2019). This study accepts that perception may always be weather-impaired, and addresses the need to consider real-time testing for downstream subsystems - without computationally expensive weather simulations, or by resorting to ad-hoc, opportunistic real-world testing as opposed to systematic, repeated scientific experiments. Key to the motivation behind this study is the idea that a truly robust AV control system can be achieved more effectively with a design where each of its component subsystems are robust to imprecision or unreliability in the other - implying *both* a reliable perception system, *and* a robust downstream processing system which is tolerant of errors in perception.

# 6 Conclusions

This study provides the following conclusions:

- Over a 20-minute period late in the day, rapidly fading light levels can have severe consequences upon perception performance - highlighting the need to develop robust subsystems for autonomous vehicle control.
- Adherent droplets cause a reduction in object detection accuracy, though the effect is significantly less pronounced than that of fading light conditions.
- Simple image augmentations can be used as a 'drop-in' component in the pipeline to apply synthetic weathering effects to both real and simulated data.
- A quantitatively accurate degradation of the perception subsystem performance can be achieved with adjustment of a single parameter for emulation of each weather condition.
- Similar parameter values may potentially be used to simulate the effects of adverse weather conditions in both augmented-reality real-world driving and simulated driving.
- The algorithms are low-latency (under 8 ms) and can be integrated into an autonomous racing vehicle running in real time on hardware that is suitable for on-board use within the vehicle.
- This augmentation method is suitable for repeatable testing in user-controlled weather conditions in both real-world and simulation - allowing the perception to be progressively worsened, and downstream subsystems developed and improved accordingly.

# Acknowledgements

The authors wish to thank: Nabil Yassine for image labelling; Peter Ball, Matthias Rolf, Gordana Collier and Fabio Cuzzolin for fruitful discussions and feedback; Jordan Trainor-Hurrell for development of simulation environments and models; Simon Jagger and Nicola Bicknell for fluid circuit design; sponsors OXTS, Datron, StreetDrone and Zeta for their continued support throughout the project.